# A NOVEL METHOD FOR THE DESIGN OF 2-DOF PARALLEL MECHANISMS FOR MACHINING APPLICATIONS

Félix Majou

*Institut de Recherches en Communications et Cybernétique de Nantes[1],*

*1 rue de la Noë, 44321 Nantes, FRANCE*

Felix.Majou@irccyn.ec-nantes.fr

Philippe Wenger

*Institut de Recherches en Communications et Cybernétique de Nantes,*

*1 rue de la Noë, 44321 Nantes, France*

Philippe.Wenger@irccyn.ec-nantes.fr

Damien Chablat

*Institut de Recherches en Communications et Cybernétique de Nantes,*

*1 rue de la Noë, 44321 Nantes, FRANCE*

Damien.Chablat@irccyn.ec-nantes.fr

**Abstract**  Parallel Kinematic Mechanisms (PKM) are interesting alternative designs for machine tools. A design method based on velocity amplification factors analysis is presented in this paper. The comparative study of two simple two-degree-of-freedom PKM dedicated to machining applications is led through this method: the common desired properties are the largest square Cartesian workspace for given kinetostatic performances. The orientation and position of the Cartesian workspace are chosen to avoid singularities and to produce the best ratio between Cartesian workspace size and mechanism size. The machine size of each resulting design is used as a comparative criterion.

**Keywords:**  Parallel Kinematic Machine Tool, Velocity Amplification Factors, Optimal Workspace Design.

## 1.     Introduction

Most industrial Machine Tools (MT) have a serial kinematic architecture: each axis supports the following one, including its actuators

---

[1] IRCCyN : UMR n° 6597 CNRS, Ecole Centrale de Nantes, Université de Nantes, Ecole des Mines de Nantes

and joints. High Speed Machining (HSM) highlights some drawbacks of such architectures: heavy moving parts require high stiffness from the machine structure to limit bending problems that lower the machine accuracy and limit the dynamic performances of the feed axes.

Parallel Kinematic Machines (PKM) attract more and more researchers and companies, because they are claimed to offer several advantages over their serial counterparts, like high structural rigidity and high dynamic capacities. Indeed, the parallel kinematic arrangement of the links provides higher stiffness and lower moving masses that reduce inertia effects. Thus, PKM have better dynamic performances, which is interesting for HSM.

However, most existing PKM have a complex geometrical workspace shape and highly non linear input/output relations. For most PKM, the Jacobian matrix which relates the joint velocities to the output velocities is not constant. Consequently, the performances may vary significantly for different points in the workspace and for different directions at one given point, which is a serious drawback for machining applications, Kim et al., 1997. To satisfy the needs of machining applications, a parallel kinematic architecture should preserve good workspace properties such as a regular shape and homogeneous kinetostatic performances throughout.

The design method presented in this paper is conducted for two-degree-of-freedom (2-DOF) mechanisms. Each mechanism is defined by a set of three design variables. The notion of useful workspace is then explained. Given prescribed kinetostatic performances, the link dimensions and actuated joint ranges of each mechanism are calculated for the largest square useful workspace. The orientation and position of the useful workspace are chosen to avoid singularities and to produce the best ratio between useful workspace and Cartesian workspace. Then, the size of the resulting mechanisms are compared.

The organization of this paper is as follows: the next section presents the kinematics of the studied mechanisms, sections 3 and 4 are devoted to the design of the two mechanisms through the velocity amplification factors analysis, and the last section concludes this paper.

## 2. Kinematic study

## 2.1 Description of the mechanisms

The two mechanisms under study have two constant length struts gliding along fixed linear actuated joints with different relative orientation. The two struts are of equal lengths *L*. Figure 1 presents the Biglide and Figure 2 shows the 2-DOF Orthoglide, Wenger et al. 2001.

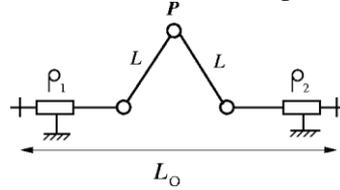

*Figure 1*. The Biglide mechanism

The joint variables are $\rho_1$ and $\rho_2$ associated with the two actuated prismatic joints and the output variables are the Cartesian coordinates of the tool center point $P = [x, y]^T$. Parameters characterizing each mechanism are the lengths $L_0$, $L$, and the actuated joint ranges $\Delta\rho$.

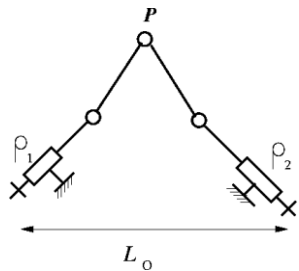

*Figure 2*. The 2-DOF Orthoglide

These two mechanisms are convenient for machining applications because they fit the technological constraints that a convenient PKM architecture for machining should respect as explained in Majou et al., 2002:
- only one DOF in each lower pair, for a simple design and a low cost;
- actuators fixed on the frame, to reduce inertia effects;
- actuated prismatic joints to allow linear motors;
- similar legs, for a low cost.

## 2.2  Velocity analysis

The Jacobian matrix relates the velocity vector $\dot{\mathbf{t}}$ of the tool point *P* to the velocity vector $\dot{\boldsymbol{\rho}}$ of the prismatic joints.

$$\dot{\boldsymbol{\rho}} = \begin{bmatrix} \dot{\rho}_1 \\ \dot{\rho}_2 \end{bmatrix} \quad \text{and} \quad \dot{\mathbf{t}} = \begin{bmatrix} \dot{x} \\ \dot{y} \end{bmatrix}$$

The vectors $\dot{\mathbf{t}}$ and $\dot{\boldsymbol{\rho}}$ are related by
$$\mathbf{A}\dot{\mathbf{t}} = \mathbf{B}\dot{\boldsymbol{\rho}}$$
where **A** and **B** are the parallel and serial Jacobian matrices, Gosselin and Angeles, 1990. When **A** and **B** are not singular, the following relations are obtained:

$$\dot{\mathbf{t}} = \mathbf{J}\dot{\boldsymbol{\rho}} \text{ with } \mathbf{J} = \mathbf{A}^{-1}\mathbf{B} \text{ and } \dot{\boldsymbol{\rho}} = \mathbf{J}^{-1}\dot{\mathbf{t}} \text{ with } \dot{\mathbf{t}} = \mathbf{J}\dot{\boldsymbol{\rho}}$$

The inverse Jacobian matrix $\mathbf{J}^{-1}$ is used for more simplicity.

## 2.3    Singularity analysis

Six types of singularities can arise in a mechanism, Zlatanov et al. 1996, but focusing on the three common ones, Gosselin and Angeles, 1990, is enough for the purpose of the work presented here.

The first type occurs when the determinant of matrix **A** vanishes, i.e. when det(**A**) = 0, Gosselin and Angeles, 1990. This type of singularity is called a parallel singularity (Fig. 3 and Fig. 4).

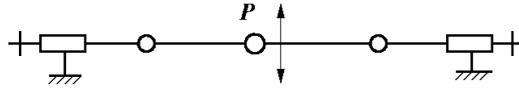

*Figure 3.* Biglide parallel singularity

In this configuration, it is possible to move the tool center point whereas the actuated joints are locked, thus the control of the tool point *P* is lost. These singularities have to be eliminated from the Cartesian workspace to prevent damaging the mechanism.

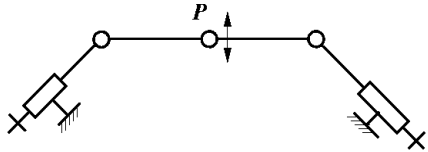

*Figure 4.* 2-DOF Orthoglide parallel singularity

The second type of singularity occurs when the determinant of matrix **B** vanishes, i.e. when det(**B**) = 0, Gosselin and Angeles, 1990. This type of singularity is called a serial singularity (Fig. 5 and Fig. 6). In this configuration, there exists a direction along which no velocity can be produced. For a PKM, serial singularities define the boundary of the Cartesian workspace, Merlet, 1997. Because the two struts are of equal lengths, the serial singularity is also a structural singularity and P can freely rotate around the two coincident revolute centers, Gosselin and Angeles, 1990.

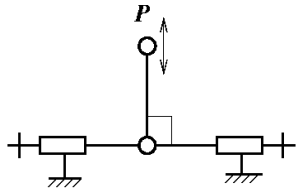 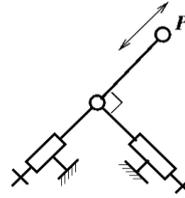

*Figure 5.* Biglide serial singularity    *Figure 6.* 2-DOF Orthoglide serial singularity

## 3. Shape, position and orientation of the useful workspace

### 3.1 Velocity amplification factor boundaries

In order to keep reasonable and homogeneous kinetostatic properties inside the Cartesian workspace, the manipulability ellipsoids of velocity defined by the inverse Jacobian matrix $\mathbf{J}^{-1}$ are studied, Yoshikawa, 1985. The $\mathbf{JJ}^{-1}$ eigenvalues square roots, $\gamma_1$ and $\gamma_2$, are the lengths of the semiaxes of the ellipse that define the two Velocity Amplification Factors (from now on called VAF) between the actuated joints velocities and the velocity vector $\dot{\mathbf{t}}$, $\lambda_1 = 1/\gamma_1$ and $\lambda_2 = 1/\gamma_2$. To limit the variations of these factors inside the Cartesian workspace, the following constraints are set

$$1/3 \leq \lambda_i \leq 3$$

This means that for a given joint velocity, the output velocity is at most three times larger or, at least, three times smaller. These constraints also permit to limit the loss of rigidity (velocity amplification lowers rigidity) and of accuracy. The boundaries on VAF were chosen as an example and should be revised depending on the machining tasks (accuracy needs for example).

### 3.2 Useful workspace shape

The Cartesian workspace (from now on called C-workspace) is the manipulator's workspace defined in Cartesian space. The useful workspace (from now on called u-workspace) is defined as a part of the C-workspace. Its shape and size are a design parameters and have to be defined. Furthermore, inside the u-workspace, VAF remain under the prescribed values.

The u-workspace shape of the two mechanisms should be similar to the one of classical serial MT, which is parallelepipedic if the machine has three translational degrees of freedom for instance. The u-workspace of a serial three axis MT is equivalent to the C-workspace because the input/output relations are linear. Therefore, a square u-workspace is prescribed here. And it must be a t-connected region, i.e. it must be free of serial and parallel singularities, Chablat and Wenger, 1998.

### 3.3  Isotropy continuum of the 2-DOF Orthoglide

The 2-DOF Orthoglide mechanism, extended to three DOF in Wenger and Chablat, 2000, was designed to have an isotropic configuration for which the VAF are unitary. But this mechanism also provides an isotropy continuum which is a straight line (Fig. 9).

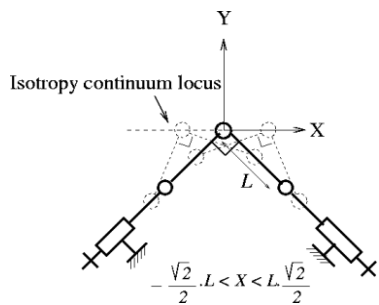 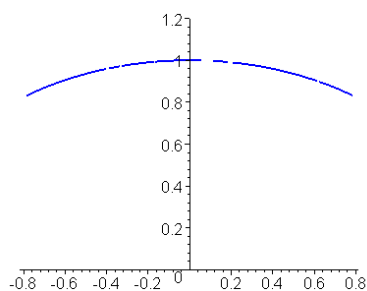

*Figure 9*. 2-DOF Orthoglide isotropy continuum locus

*Figure 10*. VAF along the isotropy continuum locus

The two VAF are equal along the continuum, but not constant, Angeles, 1997. It means that $\lambda_1 = \lambda_2$, and therefore cond(**J**) = 1. The variation of the VAF along the isotropy continuum is limited (Fig. 10), which is interesting as it shows that isotropy brings homogeneousness to kinetostatic performances, which is prefered for this application.

The Biglide isotropy continuum is not studied here because it has few consequences on the VAF homogeneousness inside the u-workspace. See section 4.1 for more details on its location.

### 3.4  Useful workspace orientation

The 2-DOF Orthoglide u-workspace is first arbitrarily centered on the point S where the VAF are equal to 1 (Fig. 11). Changing the u-workspace center position will be discussed in section 4. Two possible u-workspace orientations are studied (Fig. 11) and it appears that orientation A has a bad ratio between the u-workspace and the C-workspace, which yields a poor machine compactness because of the larger joint ranges.

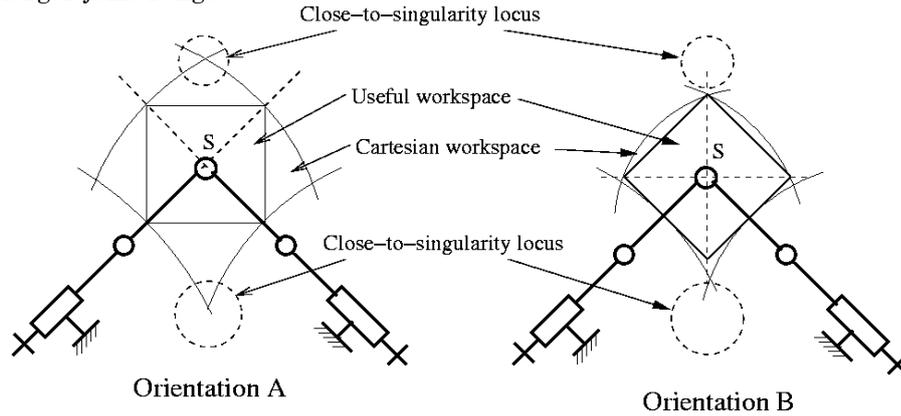

*Figure 11.* Two orientations for the 2-DOF Orthoglide u-workspace

Furthermore in the case of orientation A, singular configurations may appear inside the C-workspace, which is not acceptable. Indeed, the u-workspace is used for the machining task, but the C-workspace can be used for changing the tool position between two machining operations. Singularities are then strictly prohibited. Thus orientation B is selected for the 2-DOF Orthoglide.

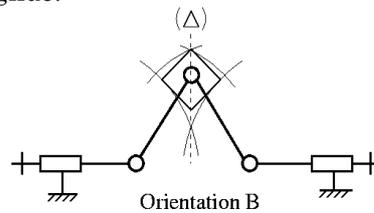

*Figure 12.* Orientation of the Biglide u-workspace

Same comments about compactness and singularities avoidance can be made for the Biglide, thus orientation B is also chosen (Fig. 12).

## 4. Optimal useful workspace design

This section explains how the u-workspace is designed: first the best workspace center locus is found by computing the VAF along the u-workspace sides. Then the u-workspace is sized so that the VAF are inside the boundaries defined in section 3.1.

## 4.1  Workspace center locus

To find the best u-workspace center locus, we shift the u-workspace perpendicularly to $(\Delta)$ and along $(\Delta)$ (Fig. 13) and the VAF are computed for each configuration.

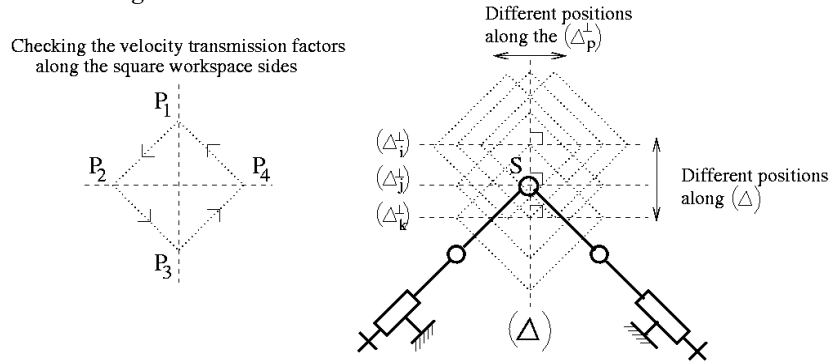

*Figure 13.* Looking for the best u-workspace center locus of 2-DOF Orthoglide

In each case, VAF extrema are located along the sides $P_iP_j$: they start from 1 at point S, then they vary until they reach prescribed boundaries on VAF (Fig. 15, section 4.2). Computing the VAF (which analytical expressions $\lambda_i$ (Xp,Yp) have been obtained with Maple) along the 4 sides of the square takes only 5 sec. with a Pentium II class PC.

As Biglide configurations are identical along every horizontal line orthogonal to $(\Delta)$ (Fig. 12), VAF are constant along these lines. Consequently, the workspace position will only be discussed along $(\Delta)$. This corresponds to the u-workspace sizing process described in section 4.2.

## 4.2  Useful workspace size

To size the u-workspace, one initial point is chosen on the locus found in section 4.1 then the u-workspace is grown until the VAF meet their limits (Fig. 14). It appears that the VAF limits are met at points P1 and P3 (Fig. 15). For the 2-DOF Orthoglide, the initial u-workspace center is point S. It appears that the first limit met is the upper one ($\lambda_i < 3$), and that it is met by $\lambda_2$, simultaneously at points P1 and P3, therefore point S remains the final u-workspace center. We see on Fig. 15 that $\lambda_1$ does not vary much compared to $\lambda_2$.

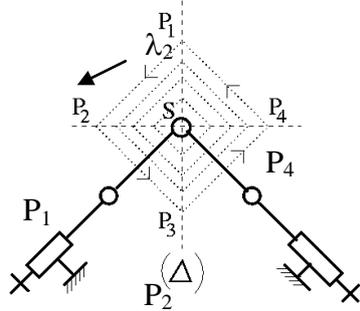 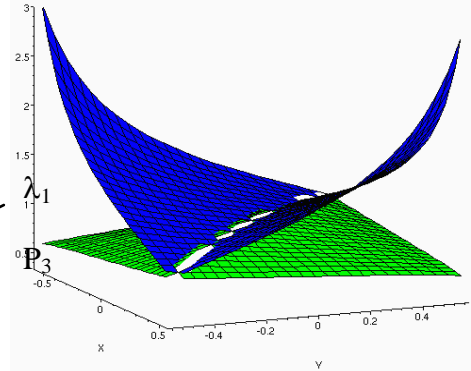

*Figure 14.* Sizing the 2-DOF Orthoglide u-workspace

*Figure 15.* VAF values of the 2-DOF Orthoglide inside its u-workspace

For the Biglide, the two boundary lines are found (Fig. 16) and the distance between them define the diagonal of the u-workspace.

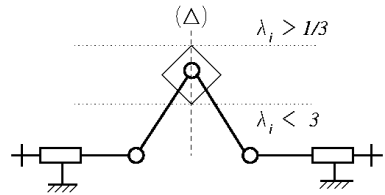

Figure 16. Sizing the Biglide u-workspace

### 4.3     Comparison of the mechanisms envelope size

For a square u-workspace of 1m², the design parameters ($L_0$, $L$, $\Delta\rho$) computed by Maple are given in Tb. 1. Obviously, the 2-DOF Orthoglide is more compact than the Biglide, Wenger et al. 2001.

*Table 1.* Design parameters for the Biglide and for the 2-DOF Orthoglide

|  | $L_0$ (m) | $L$ (m) | $\Delta\rho$ (m) | Mechanism envelope (m²) |
|---|---|---|---|---|
| Biglide | 5.95 | 3.05 | 1.67 | 16.45 |
| 2-DOF Orthoglide | 2.08 | 1.06 | 1.18 | 3.91 |

### 5.     Conclusions

The design of two 2-DOF PKM dedicated to machining applications has been conducted in this paper, through a novel design method based on the analysis of VAF. The procedure applied is reminded here: first, boundaries on VAF and u-workspace shape and size have to be defined depending on the application to achieve. Secondly, the u-workspace orientation and position have to be found inside the C-workspace for the largest ratio between u-workspace and C-workspace. Then the u-workspace is grown in the found configuration until boundaries are met.

In the case studied here, the u-workspace is square and the boundaries on VAF are 1/3 and 3. The orientation and position of the u-workspace have also been chosen to avoid singularities inside the C-workspace and to achieve best compactness. The machine size of each resulting design is used as a comparative criterion and the 2-DOF Orthoglide appeared to have smaller dimensions than the Biglide.

Further comparisons between the mechanisms studied in this paper could deal with the way VAF vary inside the u-workspace.